\documentclass[letterpaper, 10 pt, conference]{ieeeconf}

\IEEEoverridecommandlockouts

\usepackage{cite}
\usepackage{amsmath,amssymb,amsfonts}
\usepackage{algorithmic}
\usepackage{graphicx}
\usepackage{textcomp}
\usepackage{xcolor}
\usepackage{siunitx}
\usepackage{float}
\usepackage[]{footmisc}
\usepackage{ar}
\usepackage{url}
\usepackage{balance}
\usepackage{nicefrac}

\newcommand{\bs}[1]{\boldsymbol{#1}}

\renewcommand{\baselinestretch}{0.826}

\def\BibTeX{{\rm B\kern-.05em{\sc i\kern-.025em b}\kern-.08em

    T\kern-.1667em\lower.7ex\hbox{E}\kern-.125emX}}

\begin{document}

\title{\LARGE \bf Embodied Intelligence for Advanced Bioinspired Microrobotics: \\ Examples and Insights \\

\thanks{The research presented in this paper was partially funded by the Washington State University (WSU) Foundation and the Palouse Club through a Cougar Cage Award to \mbox{N.\,O.\,P.-A.}; the US National Science Foundation (NSF) through \mbox{Award\,\,2244082}; and, the Morgan Family Charitable Trust through a direct gift to \mbox{N.\,O.\,P.-A.} Additional funding was provided by the WSU Voiland College of Engineering and Architecture through a \mbox{start-up} fund to \mbox{N.\,O.\,P.-A.}.} %
\thanks{The author is with the School of Mechanical and Materials Engineering, Washington State University (WSU), Pullman,\,WA\,99164,\,USA. \mbox{E-mail:} {\tt n.perezarancibia@wsu.edu}.}%
%}
}
\author{N\'estor\,O.\,P\'erez-Arancibia}

\maketitle
\thispagestyle{empty}
\pagestyle{empty}

\begin{abstract}
The term \textit{embodied intelligence} (EI) conveys the notion that body morphology, material properties, interaction with the environment, and control strategies can be purposefully integrated into the process of robotic design to generate intelligent behavior; in particular, locomotion and navigation. In this paper, we discuss EI as a design principle for advanced microrobotics, with a particular focus on \mbox{co-design}---the simultaneous and interdependent development of physical structure and behavioral function. To illustrate the contrast between \mbox{EI-inspired} systems and traditional architectures that decouple sensing, computation, and actuation, we present and discuss a collection of robots developed by the author and his team at the \textit{Autonomous Microrobotic Systems Laboratory} (AMSL). These robots exhibit intelligent behavior that emerges from their structural dynamics and the physical interaction between their components and with the environment. Platforms such as the \mbox{Bee\textsuperscript{++}}, \mbox{RoBeetle}, \mbox{SMALLBug}, \mbox{SMARTI}, \mbox{WaterStrider}, \mbox{VLEIBot\textsuperscript{+}}, and \mbox{FRISSHBot} exemplify how feedback loops, decision logics, sensing mechanisms, and smart actuation strategies can be embedded into the physical properties of the robotic system itself. Along these lines, we contend that \mbox{co-design} is not only a method for empirical optimization under constraints, but also an enabler of EI, offering a scalable and robust alternative to classical control for robotics at the \mbox{mm-to-cm--scale}.
\end{abstract}

\section{Introduction}
\vspace{-0.5ex}
\label{SEC01}
Traditionally, in the fields of robotics and modern control, sensing, perception, \mbox{decision-making}, and actuation are treated as distinct modules implemented and connected through electronics for communication, signal processing, and control. This architecture---ubiquitous from industrial manipulators to autonomous vehicles---relies heavily on centralized programmable controllers and logics. In contrast, the concept of \textit{embodied intelligence} (EI) is based on the notion that complex behavior can emerge from the physical interaction between a robot’s structural components, its materials, and its environment. According to this analytical framework, morphology and materials are not merely physical properties of a robot's body, but active elements in the processes of sensing, actuation, computation, and control. Automated motion patterns and feedback for functionality, stability, and, foreseeably, for \mbox{decision-making} can be \textit{encoded} in the structure and dynamics of the physical system itself.

The basic principles underlying EI---namely, the integration of sensing, actuation, computation, and control into the physical structure of a system---can be conceptually traced back to ancient mechanical devices. One of the earliest known examples is Ktesibios' clepsydra (water clock), which used \mbox{float-regulated} valves to maintain a constant water level, thus embodying feedback control through direct material interaction with the environment~\cite{Drachmann1948}. Similar principles can be identified in the automata developed by Hero of Alexandria, who is believed to have designed \mbox{steam-driven} and \mbox{air-powered} devices capable of performing complex sequences of actions through intelligent use of mechanical linkages and timing mechanisms~\cite{Hero1851}. During the \mbox{so-called} Islamic Golden Age, \mbox{Al-Jazari} built programmable automata and water clocks that incorporated float valves, cams, and timing mechanisms---advancing the integration of structure and behavior~\cite{AlJazari1974}. 

More recently, in the early 19th century, \mbox{J.\,M.\,Jacquard} introduced a system of punch cards to control weaving patterns, which represents an early fusion of mechanical structure with encoded behavior~\cite{Essinger2004}. Perhaps, the most \mbox{well-known} system exhibiting some key features of EI---such as electronics-free control---is \mbox{J.\,Watt's} centrifugal governor, which additionally is considered to be the most direct antecedent to modern feedback control systems. This mechanism was widely used in Victorian England to regulate the speeds of steam engines via a purely mechanical feedback loop that tightly coupled actuation and sensing~\cite{Maxwell1867,Bennett1979}. Furthermore, this system inspired \mbox{J.\,C.\,Maxwell} to develop the first modern method for analyzing the stability of \mbox{feedback-controlled} \mbox{closed-loop} systems. Collectively, these examples illustrate how EI, though formally articulated only recently, has deep historical roots in mechanical design.
\begin{figure*}[t!]
\vspace{1.6ex}    
\begin{center}    
\includegraphics[width=0.98\textwidth]{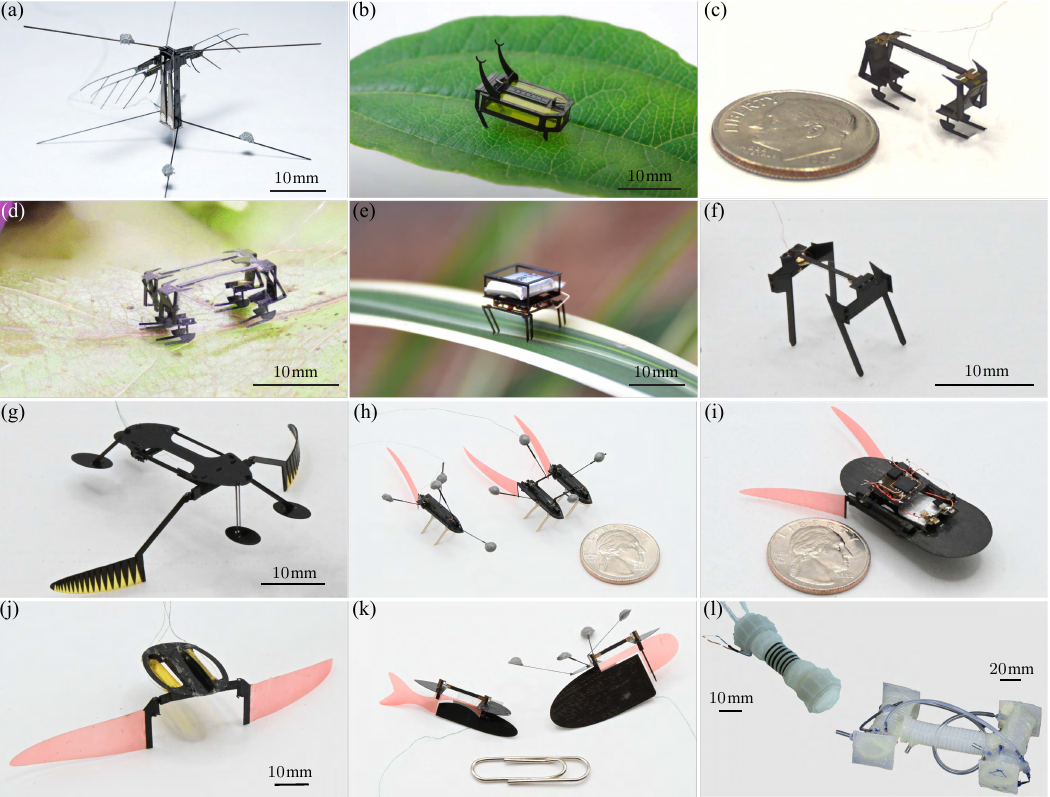}
\end{center}
\vspace{-2ex}
\caption{\textbf{Embodied intelligence in action.} All the microrobots presented in this figure were developed by the author and his team at the AMSL, and each exhibits some form of EI at some level in its design.~\textbf{(a)}\,The \mbox{$95$-mg} Bee\textsuperscript{++}, which leverages FSI and an \mbox{ISP-based} mechanism in combination with modern nonlinear \mbox{Lyapunov-based} methods, to control its yaw DOF.~\textbf{(b)}\,The \mbox{$88$-mg} \mbox{RoBeetle}, an autonomous \mbox{anisotropic-friction-based} crawler mechanically powered by an \mbox{NiTi--Pt} composite catalytic artificial muscle. This muscle uses the flameless catalytic combustion of methanol, enabled by a rough layer of Pt, to thermally excite its core made of \mbox{NiTi} SMA in cycles that produce periodic actuation. The system that controls the \mbox{catalytic-combustion} process and the phase transitions of the SMA material is entirely \mbox{electronics-free}, thus leveraging EI.~\textbf{(c)}\,The SMALLBug, a crawler that uses a \mbox{high-frequency} \mbox{SMA-based} bending actuator to locomote on flat surfaces. In this design, the cyclic bending motion of the driving actuator is transformed by a \mbox{$2\Sigma$-shaped} frame into rectilinear locomotion through leveraging anisotropic friction.~\textbf{(d)}\,The SMARTI, a crawler composed of two SMALLBug platforms connected in parallel. This configuration is \mbox{$2$D} steerable simply by exciting its two driving actuators with \mbox{phase-shifted} PWM voltages, thus leveraging anisotropic friction for functionality and control.~\textbf{(e)}\,The SPARQ, a fully autonomous---from both the power and control perspectives---crawler, which was conceived as an advanced version of the SMARTI.~\textbf{(f)}\,The MiniBug, which, at \mbox{10\,mg}, is the smallest and lightest crawler with onboard actuation ever created. This robot was developed by miniaturizing the actuator and frame of the \mbox{SMALLBug}, and by adopting the legs of the SPARQ.~\textbf{(g)}\,The WaterStrider, a surface swimmer that uses \mbox{rowing-resembling} motion patterns of two paddles and anisotropic drag to propel itself forward.~\textbf{(h)}\,The VLEIBot (left) and VLEIBot\textsuperscript{+} (right) are two robots that use propulsors inspired by anguilliforms to swim forward and steer themselves. These propulsors generate thrust by leveraging FSI as traveling waves are induced by flapping soft passive fins with \mbox{SMA-based} bending actuators.~\textbf{(i)}\,The VLEIBot\textsuperscript{++}, a fully autonomous---from both the power and control perspectives---swimmer, which was conceived as an advanced version of the VLEIBot\textsuperscript{+}.~\textbf{(j)}\,The BILLEBot, a swimmer design that combines the transmission mechanism of the WaterStrider with the \mbox{traveling-wave-based} \mbox{thrust-generation} method of the VLEIBot platform.~\textbf{(k)}\,The old FRISSHBot (left) and new FRISSHBot (right) are two swimmers composed of two plates connected by an \mbox{SMA-based} bending actuator that applies periodic torques---with equal magnitudes and opposite directions---to both of them during operation. These actuation torques induce hydrodynamic reactive torques---generated by aggregated inertial and viscous forces---on the plates that, as a consequence, produce the thrust required for swimming. Specifically, by design, the front plate functions as an anchor and the rear plate as a caudal flapping rigid fin. This flapping fin, through the creation of a couple of vortices during an operation cycle---one clockwise and the other counterclockwise---induces a jet that, by conservation of momentum, propels the robot forward.~\textbf{(l)}\,The pneumatic soft robot on the left can locomote inside pipes and trenches by leveraging \mbox{pressure-controlled} anisotropic friction, using a distributed stretchable artificial skin as the main sensor. The pneumatic soft robot on the right can locomote inside pipes and trenches driven by an entirely \mbox{electronics-free} feedback controller based on neuromorphic mechanical computation, using mechanical tactile and proprioceptive sensors. \label{FIG01}}
\vspace{-1.0ex}
\end{figure*}

The formal notion of EI was first formulated in the early 1990s, primarily as a response to the limitations of symbolic \textit{artificial intelligence} (AI). A key new idea was introduced by \mbox{R.\,A.\,Brooks}, who argued that intelligent behavior does not necessarily require internal representations but can in some instances arise from direct sensorimotor interaction with the \mbox{environment~\cite{Brooks1991}}. His work on \mbox{behavior-based} robotics explored the notion that intelligence is grounded in physical embodiment and environmental coupling. Building upon these developments, \mbox{R.\,Pfeifer} and \mbox{C.\,Scheier} formulated a comprehensive theoretical framework that emphasizes the computational role of morphology, materials, and \mbox{real-time} environmental interaction in shaping \mbox{behavior~\cite{Pfeifer1999}}. Specifically, regarding the main topic of this article, they state that \textit{``a fundamental consequence of embodiment is that embodied agents must interact with their environments. To understand this interaction, we have to study, for
example, how organisms acquire experience: knowledge about
the environment obtained by interacting with it.''} 

An earlier theoretical biological foundation for EI lies in the work of \mbox{H.\,Maturana} and \mbox{F.\,Varela}, who coined the concept of \textit{autopoiesis} to describe the \mbox{self-producing} nature of living \mbox{systems~\cite{Maturana1980}}. In their formulation, cognition is not simply the processing of abstract symbols or representations but is instead rooted in the dynamic and recursive interactions between an organism and its environment. This view was formalized through the idea of \textit{structural coupling}, whereby a living system maintains its identity while continuously adapting its internal organization through sensorimotor engagement with its surroundings. Maturana~and~Varela claim that \textit{``living is cognition,''} which implies that intelligent behavior emerges from the embodied \mbox{self-maintaining} dynamics of the organism itself. This biological perspective significantly influenced later research on embodied robotics and enactive cognitive science, fields in which intelligence is conceived as an emergent property of \mbox{real-time} physical interaction with the world---rather than the result of disembodied computation or centralized control architectures.

In this paper---using as examples several robots developed by the author and his team at the \textit{Autonomous Microrobotic Systems Laboratory} (AMSL)~\cite{YangX2019,BenaRM2022,BenaRM2023,YangX2020,NguyenXT2020,BenaRM2021,TrygstadCK2023,BlankenshipEK2024,LongwellCR2024,TrygstadCK2025I,TrygstadCK2024, TrygstadCK2025II,XuK2020}---we contend that the basic concepts of EI, which at first sight might appear complex and abstract, can be adopted to guide the design and development of bioinspired microrobots. Through these examples, we present \mbox{co-design}---as opposed to modular design---as a central principle of EI, especially to address problems relating to power, computation, and control imposed by the stringent constraints of volume and weight inherent to systems operating at the \mbox{mm-to-cm--scale}.

The rest of the paper is organized as follows. \mbox{Section\,\ref{SEC02}} discusses the main differences between classical control for robotics and EI in enabling complex behaviors in locomoting robots. \mbox{Section\,\ref{SEC03}} presents several examples of EI in action through the description of key features of microrobots developed by the author and his team during the past few years. Last, \mbox{Section\,\ref{SEC04}} summarizes the ideas presented in the paper. 

\section{Classical Control vs. Embodied Intelligence}
\vspace{-0.5ex}
\label{SEC02}
In classical control for robotics, a standard \mbox{closed-loop} system is composed of~(i)\,sensors that perceive the surrounding environment and measure the state of the system;\,(ii)\,a controller that processes the information provided by the sensors, using algorithms based on mathematical laws and logic, to generate the commands that excite the system's actuators; and,\,(iii)\,the actuators that generate the forces and torques necessary to interact intelligently with the environment, according to the commands computed by the controller. This separation of perception (sensing), computation (control), and actuation requires explicit feedback, state estimation, and decision algorithms, which are predominantly implemented through the use of electronics. EI, by contrast, integrates some or all of these functions physically into the robot. For instance, morphology---including structure, compliance, and geometry---material dynamics, and control strategies can be purposely \mbox{co-designed} to interact with the environment so that intelligent behavior emerges through the physics of the system and its interaction with the environment.

As already hinted, here we refer to a \textit{controller} not necessarily as an \mbox{electronics-based} device---analog or digital---but, possibly, to a \mbox{morphology-material-environment} loop. The physical realization of this concept is clearly observable in the robots presented in~\cite{YangX2020}~and~\cite{XuK2020}. From these cases, it can be inferred that the distinction between classical control and EI leads to a different \mbox{robotic-design} paradigm and a different set of development tools. Classical controllers are \mbox{well-suited} for driving highly deterministic robotic systems that operate with low levels of uncertainty---e.g., in car or semiconductor factories---where flexibility, modularity, programmability, and high precision can be achieved. Conversely, \mbox{EI-based} design can be employed to prioritize robustness, efficiency, and simplicity by using less \mbox{general-purpose} architectures. These \mbox{less-general} architectures, however, can be optimized for specific tasks under severe constraints. For example, EI is particularly well suited for microrobotics design because, as robots are scaled down in size and mass, conventional sources of energy and architectures become infeasible due to limitations in battery storage, computational power, and sensor integration. We believe that, at small scales, it is reasonable to design systems in which control emerges---entirely or partially---from physical design rather than exclusively from abstract computation.

Generally, to systematize EI for microrobotics development, a design framework must account for the nonlinear, often nonintuitive interactions among structure, materials, and behavior. Here, we propose a small set of heuristic \mbox{EI-based} design guidelines:
\begin{itemize}
\item[i.] Identify the ambient physical phenomena that can be leveraged for actuation, sensing, and control purposes---for example, anisotropic friction, anisotropic drag, buoyancy, and thermal lag.
\item[ii.] Determine the morphological features that amplify or modulate the physical phenomena leveraged for actuation, sensing, and control.
\item[iii.] Use analyses, numerical simulations, and experiments to understand the dynamics of the actuators that map \textit{energy} inputs into \textit{behavioral} outputs.
\item[iv.] \mbox{Co-design} morphology and actuation to create and stabilize desired behaviors.
\item[v.] Accept limited or more complex programmability as a tradeoff for robustness, simplicity, and weight.
\item[vi.] Observe living systems---particularly, animals and plants---for inspiration, as intelligent behavior emerges from the embodied \mbox{self-maintaining} dynamics of the organisms themselves.
\end{itemize}

Next, we illustrate these ideas using some of the robots developed by the author and his team at the AMSL. 

\section{Embodied Intelligence in Action}
\vspace{-0.5ex}
\label{SEC03}
In this section, we discuss how the principles of EI---which at first sight may seem complex and ethereal---can be used to solve problems in microrobotics, in which stringent constraints relating to mass, volume, power, control, computation, sensing, and actuation must be addressed. Here, we accomplish that objective through the presentation of examples; namely, microrobots developed by the author and his team during the past six years. 
\begin{figure*}[t!]
\vspace{1.6ex}    
\begin{center}    
\includegraphics[width=0.99\textwidth]{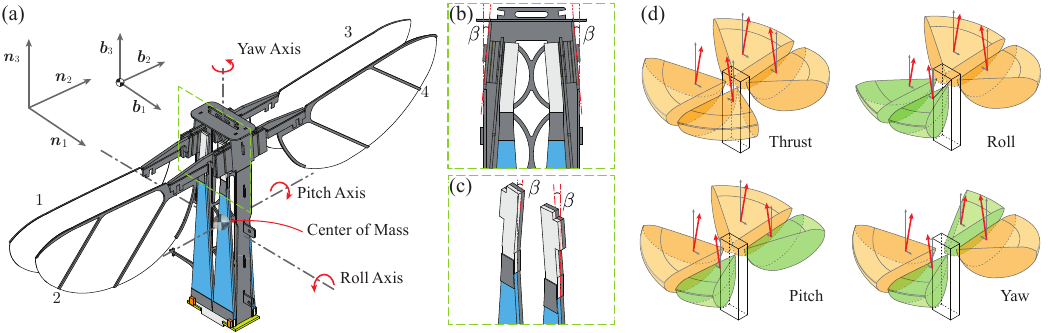}
\end{center}
\vspace{-2ex}
\caption{\textbf{The Bee\textsuperscript{++} and its flapping modes.}~\textbf{(a)}\,The \mbox{$95$-mg} Bee\textsuperscript{++} is the first \mbox{four-actuator} \mbox{four-wing} robotic insect ever developed. As discussed in~\cite{BenaRM2023}, this flyer is \mbox{six-DOF} controllable during flight.~\textbf{(b)}\,Each wing is connected to its respective unimorph actuator through a \mbox{four-bar} transmission mechanism installed with an inclination \mbox{$\beta$\,rad} relative to the \mbox{$\bs{b}_2$-$\bs{b}_3$} plane---equivalent to an inclination \mbox{$\left(\pi/2 - \beta\right)$\,rad} relative to the \mbox{$\bs{b}_1$-$\bs{b}_2$} plane.~\textbf{(c)}\,The \mbox{$\beta$-inclined} installation of each transmission of the robot is the result of a feature of the actuator design, whose connecting element at its distal end is tilted by the angle $\beta$.~\textbf{(d)}\,The Bee\textsuperscript{++} is underactuated as it has fewer actuators than degrees of freedom; however, it can be controlled using four basic flapping modes and nonlinear Lyapunov methods. The first mode consists of flapping the four wings of the robot with identical frequencies and amplitudes; thrust can be modulated by varying the amplitude of flapping during flight. The second mode consists of flapping the wings of the robot in pairs and asymmetrically with respect to the \mbox{$\bs{b}_1$-$\bs{b}_3$} plane; the roll DOF can be modulated by increasing---or decreasing---the amplitudes of flapping of one pair of wings relative to those of the other pair. The third mode consists of flapping the wings of the robot in pairs and asymmetrically with respect to the \mbox{$\bs{b}_2$-$\bs{b}_3$} plane; the pitch DOF can be modulated by increasing---or decreasing---the amplitudes of flapping of one pair of wings relative to those of the other pair. The fourth mode consists of flapping the wings of the robot in diagonal pairs; the yaw DOF can be modulated by increasing---or decreasing---the amplitudes of flapping of one pair of wings relative to those of the other pair. Note that the inclination of the stroke plane relative to the \mbox{$\bs{b}_1$-$\bs{b}_2$} plane generates a yaw torque that can be modulated to actuate and control the yaw DOF, thus exemplifying EI in action. \label{FIG02}}
\vspace{2ex}    
\begin{center}    
\includegraphics[width=0.99\textwidth]{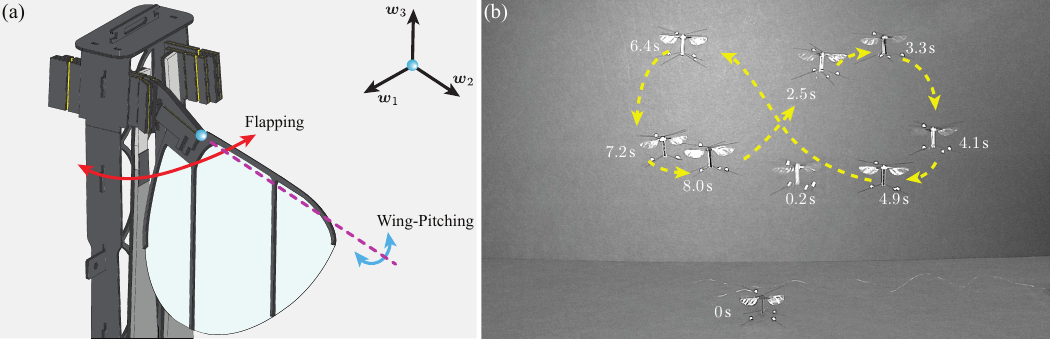}
\end{center}
\vspace{-2ex}
\caption{\textbf{Wing motion during flight and aerobatic maneuver.}~\textbf{(a)}\,Active flapping and passive \mbox{wing-pitching}. The flapping motion is generated through a \mbox{four-bar} mechanism transmission by a unimorph actuator; in contrast, the \mbox{wing-pitching} motion is the result of FSI. Combined with the \mbox{inclined-stroke-plane--based} mechanism for yaw control, the passive pitching of the wing exemplifies a case of EI in action.~\textbf{(b)}\,A Bee\textsuperscript{++} prototype during aerobatic flight. In this case, the yaw DOF was controlled by combining the \mbox{ISP-based} mechanism with nonlinear \mbox{Lyapunov-based} control schemes. This case demonstrates that EI and modern control are not incompatible and can be employed in conjunction. \label{FIG03}}
\vspace{-2.0ex}
\end{figure*}

\subsection{The Bee\textsuperscript{++}: Inclined Stroke Planes for Yaw Control}
\vspace{-0.5ex}
\label{SEC03A}
The \mbox{$95$-mg} flying Bee\textsuperscript{++} presented~in~\cite{YangX2019}~and~\cite{BenaRM2022} is an evolved version of the Bee\textsuperscript{+} first presented in~\cite{BenaRM2023}. The latter is the first \mbox{four-wing} \mbox{four-actuator} \mbox{flapping-wing} \mbox{insect-scale} flyer ever reported, and its development was enabled by the invention of a new type of \mbox{high-performance} \mbox{ultra-light} (\mbox{$8.4$\,mg}) unimorph piezoelectric actuator. A photograph of a Bee\textsuperscript{++} prototype is shown in \mbox{Fig.\,\ref{FIG01}(a)}. The design and functionality of this prototype are depicted in \mbox{Fig.\,\ref{FIG02}(a)}. In this illustration, the vector triplets $\left\{\bs{n}_1,\bs{n}_2,\bs{n}_3\right\}$ and $\left\{\bs{b}_1,\bs{b}_2,\bs{b}_3\right\}$ denote the \mbox{earth-fixed} inertial and \mbox{body-fixed} frames of reference. The robot's roll \textit{degree of freedom} (DOF) is defined as a rotation about $\bs{b}_1$; accordingly, pitching and yawing are defined as rotations about $\bs{b}_2$ and $\bs{b}_3$, respectively. Each of the four wings of the robot is driven by a single unimorph actuator---depicted in blue---through a \mbox{four-bar} mechanism of the type presented in~\cite{WoodRJ2008}. As indicated in \mbox{Figs.\,\ref{FIG02}(b)}~and~(c), the actuators are connected to their respective transmissions with an angle $\beta$, such that the stroke plane of the flapping motion is inclined relative to the \mbox{$\bs{b}_1$-$\bs{b}_2$} plane. The four flapping modes that enable the modulation of thrust and the roll, pitch, and yaw torques are shown in~\mbox{Fig.\,\ref{FIG02}(d)}. 

The motion of a Bee\textsuperscript{++} prototype's wing can be decomposed into flapping---in direction $\bs{w}_3$, which coincides with $\bs{b}_3$---and \mbox{wing-pitching}---in direction $\bs{w}_2$, as seen in~\mbox{Fig.\,\ref{FIG03}(a)}. As explained above, the flapping motion of each wing is \textit{actively} driven by an actuator; therefore, it can be modulated and controlled as described in~\cite{YangX2019,BenaRM2022,BenaRM2023}. The pitching motion of each wing is passively generated as a flexure hinge deflects as the result of the interaction of the wing with the surrounding fluid during a flapping cycle. In this way, the thrust forces required for flight and control are produced through nonlinear \textit{\mbox{fluid--structure} interaction} (FSI), a complex process very difficult to model and control explicitly. Yet, this is a clear example of morphology, actuation, and control \mbox{co-design} in which the ability to fly emerges as a manifestation of EI. The most notable characteristic of the Bee\textsuperscript{++} design, however, is its ability to fly under \mbox{six-DOF} control using the \mbox{inclined-stroke-plane--based} mechanism depicted in \mbox{Figs.\,\ref{FIG02}(b)~and~(c)}. Specifically, this approach allowed us to robustly control the yaw DOF, a \mbox{long-standing} problem in \mbox{bee-inspired} microrobotics. As depicted in the \mbox{right-bottom} illustration of \mbox{Fig.\,\ref{FIG02}(d)}, by flapping the wings of the robot in diagonal pairs of identical amplitude, a pair of \mbox{cyclic-averaged} force components---with equal magnitudes and opposite directions---lie in the \mbox{$\bs{b}_1$-$\bs{b}_2$} plane at a distance from the axis $\bs{b}_3$, thereby generating a yaw torque that can be modulated and controlled in real time. As first presented in~\cite{BenaRM2023}, at a frequency of \mbox{$165$\,Hz} and using \mbox{Lyapunov-based} nonlinear methods, sustained \mbox{high-performance} aerobatic \mbox{six-DOF-controlled} flight can be achieved, as shown in~\mbox{Fig.\,\ref{FIG03}(b)}. In this manner, by leveraging robotic, aerodynamic, and controller \mbox{co-design}---once more inspired by the notion of EI---we achieved the highest flight performance of a \mbox{bee-inspired} robot ever reported at the time of publication. Video footage of a flight test of this type is shown in the accompanying supplementary movie.
\begin{figure*}[t!]
\vspace{1.6ex}    
\begin{center}    
\includegraphics[width=0.99\textwidth]{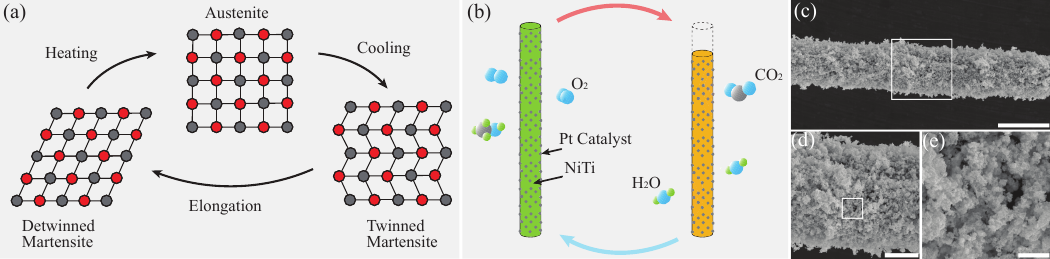}
\end{center}
\vspace{-2ex}
\caption{\textbf{Functionality of the catalytic artificial muscle developed to drive the \mbox{RoBeetle} shown in \mbox{Fig.\,\ref{FIG01}(b)}.}~\textbf{(a)}\,During a full actuation cycle, the material of an SMA wire transitions between three distinct \mbox{crystal-structure} states: \textit{detwinned martensite}, \textit{austenite}, and \textit{twinned martensite} by the sequential application of heat, allowance for convective cooling, and application of stress.~\textbf{(b)}\,In the case of the RoBeetle's muscle, controlled heat is applied through the catalytic combustion of methanol---facilitated by a rough layer of Pt, a multipurpose catalyst---and stress is applied using a CF leaf spring (shown in \mbox{Fig.\,\ref{FIG05}(b)}). Accordingly, an actuation cycle starts at the detwinned martensite state---corresponding to the extended state of the wire at room temperature; then, after surpassing the transition temperature of the SMA material, the system transitions to the austenite state---corresponding to the contracted state of the wire; next, after cooling down through free convection and reaching room temperature, the system transitions to the martensite state. The stress applied by the leaf spring to the wire continually detwins the SMA material.~\mbox{\textbf{(c)}--\textbf{(e)}}\,SEM images of the surface of an \mbox{NiTi-Pt} composite wire with a diameter of \mbox{$87$\,{\textmu}m}. The rough and porous catalytic layer (\mbox{Pt-black}) has a thickness of \mbox{$18.1$\,{\textmu}m}. The magnifications of the images are ${\times}350$, ${\times}1\hspace{0.2ex}200$, and ${\times}6\hspace{0.2ex}500$; the scale bars indicate distances of $100$, $30$, and \mbox{$5$\,{\textmu}m}, respectively. \label{FIG04}}
\vspace{2.0ex}   
\begin{center}    
\includegraphics[width=0.99\textwidth]{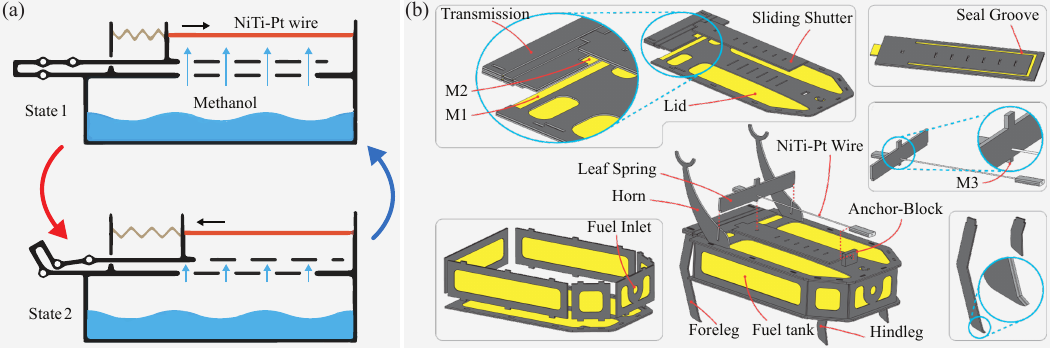}
\end{center}
\vspace{-2ex}
\caption{\textbf{Design and functionality of the RoBeetle platform.}~\textbf{(a)}\,Mechanical system that controls the \mbox{catalytic-combustion} process and the phase transitions of the muscle's SMA material without electronics, thereby leveraging EI; the robot's body simultaneously functions as an actuator, a sensor, and a mechanical controller.~\textbf{(b)}\,The RoBeetle and its functional components. In this illustration, the labels M1, M2, and M3 denote mating points of assembly fixtures.\label{FIG05}}
\vspace{-2.0ex}
\end{figure*}

\subsection{The RoBeetle: A Crawler Driven by an Artificial Muscle Controlled Using an Electronics-Free Mechanism}
\vspace{-0.5ex}
\label{SEC03B}
The RoBeetle shown in \mbox{Fig.\,\ref{FIG01}(b)}---first presented in~\cite{YangX2020}---is an \mbox{$88$-mg} autonomous crawler that leverages anisotropic friction to locomote and is actuated by a catalytic artificial muscle. The functionality of this actuator is depicted in \mbox{Fig.\,\ref{FIG04}}. The muscle has a thin cylindrical core (a wire) made of \textit{shape-memory alloy} (SMA) nitinol and a rough granular layer made of platinum (Pt) particles. An SMA wire under tension can be thermally excited according to homogeneous cycles of heating and cooling to generate periodic actuation, as the wire cyclically contracts and extends. This phenomenon is graphically explained in~\mbox{Fig.\,\ref{FIG04}(a)}, which depicts an actuation cycle that starts with the SMA wire extended because its SMA material is in a \textit{detwinned martensite} state at room temperature (\mbox{$20$\hspace{0.2ex}\textdegree{C}}). Then, through the application of heat, the SMA material transitions to \textit{austenite} and, as a consequence, the wire contracts. When the material is left to cool down by free convection, it transitions to a \textit{twinned martensite} state. Last in an actuation cycle, the application of stress detwins the SMA material, and the wire returns to its initial condition. In the case of the muscle actuating the RoBeetle, heat is applied cyclically using catalytic combustion, as shown in \mbox{Fig.\,\ref{FIG04}(b)}. This method leverages the catalytic capability of Pt, which facilitates the flameless combustion of a wide variety of fuels, including hydrogen, propane, butane, and methanol. Given that the nominal transition temperature of the SMA wire (Dynalloy NiTi with a diameter of \mbox{$50.8$\,{\textmu}m}) driving the RoBeetle is \mbox{$90$\hspace{0.2ex}\textdegree{C}}, we selected methanol as the source of energy because it ignites catalytically---using Pt as the catalyst---at temperatures as low as \mbox{$40$\hspace{0.2ex}\textdegree{C}} and can be stored easily in liquid form. Furthermore, the specific energy of methanol is on the order of \mbox{$20$\,MJ/kg}, which is substantially higher than that of the best \mbox{state-of-the-art} \mbox{Lithium-ion} (\mbox{Li-ion}) batteries available today, which ranges from $0.9$ to \mbox{$1.3$\,MJ/kg}.  

\begin{figure}[t!]
\vspace{1.6ex}    
\begin{center}    
\includegraphics[width=0.48\textwidth]{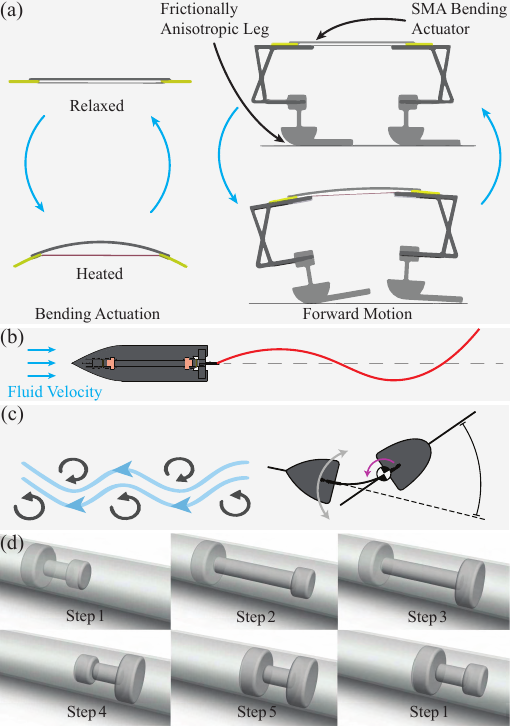}
\end{center}
\vspace{-2ex}
\caption{\textbf{Mechanisms of locomotion, leveraging EI, of some of the robots discussed in this article.}~\textbf{(a)}\,Bending actuator (left) and \mbox{anisotropic-friction-based} locomotion mode of the \mbox{SMALLBug} platform shown in~\mbox{Fig.\,\ref{FIG01}(c)}.~\textbf{(b)}\,Swimming thrust generated by a traveling wave passing through the unactuated soft slender tail of a \mbox{VLEIBot} prototype shown in~\mbox{Fig.\,\ref{FIG01}(h)}.~\textbf{(c)}\,Swimming thrust generated by \mbox{tail-generated} vortices used by the FRISSHBot platforms, shown in~\mbox{Fig.\,\ref{FIG01}(k)}, to propel themselves forward.~\textbf{(d)}\,Basic locomotion mechanism used by the pneumatic soft robots shown in~\mbox{Fig.\,\ref{FIG01}(l)}. \label{FIG06}}
\vspace{-2ex}
\end{figure}

Three images obtained using \textit{scanning electron microscopy} (SEM)---with three levels of magnification---are presented in~\mbox{Figs.\,\ref{FIG04}(c)--(e)}. We discovered experimentally that a highly granular and rough structure of this type is required to initiate and sustain catalytic combustion for actuation. In principle, the idea of using a catalytic muscle of the type shown in \mbox{Fig.\,\ref{FIG04}(b)} to mechanically power a subgram \mbox{mm-scale} crawler is straightforward. However, to implement this idea, we need to control two complex systems simultaneously under extremely stringent constraints of power, volume, and mass. These constraints greatly limit the possibility of using sophisticated sensing, electronics, and computation according to the classical control for robotics paradigm discussed in \mbox{Section\,\ref{SEC02}}. To solve this problem, we designed a system that controls the \mbox{catalytic-combustion} process and the phase transitions of the muscle’s SMA material without electronics, thereby leveraging EI; the robot’s body simultaneously functions as an actuator, a sensor, and a mechanical controller, as depicted in \mbox{Figs.\,\ref{FIG05}(a)}~and~(b). In summary, the RoBeetle's design embodies intelligence by controlling gait dynamics using structural stiffness and unactuated valving, enabling forward locomotion without any electronic feedback. Gait control emerges from the interplay of structural elasticity, actuator timing, and \mbox{surface--contact} friction. The electrically powered SMALLBug platform shown in \mbox{Fig.\,\ref{FIG01}(c)} employs a similar mode of locomotion, as shown in \mbox{Fig.\,\ref{FIG06}(a)}. A photographic composite of a RoBeetle prototype crawling on a flat surface is shown in~\mbox{Fig.\,\ref{FIG07}(a)}. Video footage of an experiment of this type is shown in the accompanying supplementary movie.

\subsection{\mbox{SMA-Based} Bending Actuation and Anisotropic Friction for Locomotion on Flat Surfaces}
\vspace{-0.5ex}
\label{SEC03C}
The SMALLBug---the crawler shown in \mbox{Fig.\,\ref{FIG01}(c)}---uses a \mbox{high-frequency} \mbox{SMA-based} bending actuator of the type shown on the left in~\mbox{Fig.\,\ref{FIG06}(a)} to locomote on flat surfaces. In this design, the two SMA wires of the actuator are connected under tension in parallel to a \textit{carbon fiber} (CF) beam, as described in~\cite{NguyenXT2020}. Accordingly, when the SMA material of the wires is cyclically Joule heated and cooled down by free convection---as explained in \mbox{Section\,\ref{SEC03B}} and depicted in \mbox{Fig.\,\ref{FIG04}(a)}---periodic actuation is produced as the SMA wires sequentially transition under tension from the detwinned martensite state to the austenite state, then to the twinned martensite state, and, last in the cycle, back to the detwinned martensite state. In this idealization, the CF beam functions as a leaf spring that facilitates the detwinning process of the SMA material. As shown on the right in~\mbox{Fig.\,\ref{FIG06}(a)}, the cyclic bending motion of the driving actuator is transformed by a \mbox{$2\Sigma$-shaped} frame into rectilinear locomotion by leveraging anisotropic friction. 

\begin{figure*}[t!]
\vspace{1.6ex}    
\begin{center}    
\includegraphics[width=0.98\textwidth]{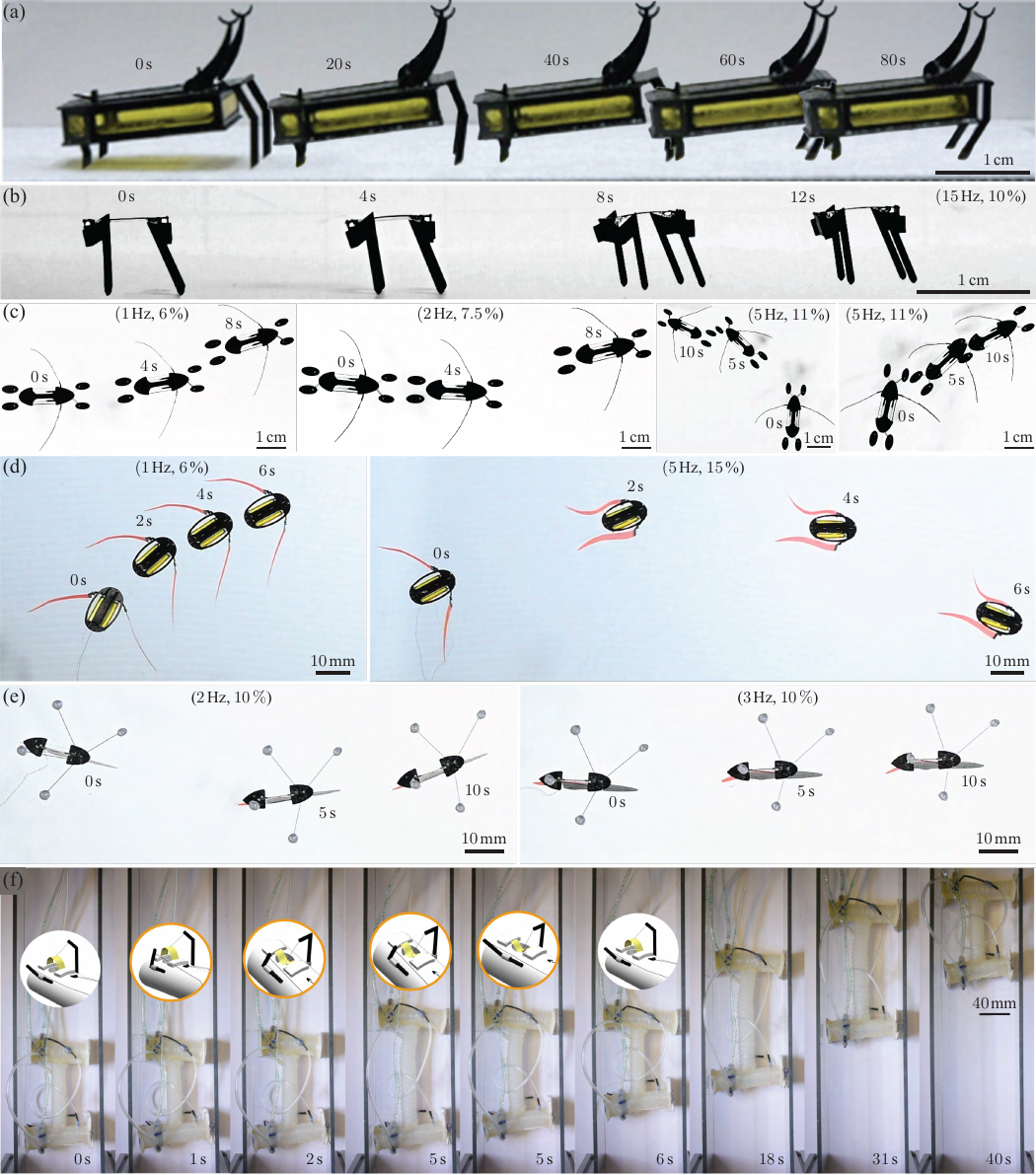}
\end{center}
\vspace{-2ex}
\caption{\textbf{Photographic composites of images showing locomotion tests of six of the robots discussed in this article.}~\textbf{(a)}\,A RoBeetle prototype crawling on a flat surface.~\textbf{(b)}\,A MiniBug prototype crawling on a flat surface, excited with a \mbox{$15$-Hz} PWM signal with a DC of \mbox{$10$\hspace{0.1ex}\%}.~\textbf{(c)}\,A WaterStrider prototype swimming excited by four different PWM signals with the frequencies and DC values in parentheses.~\textbf{(d)}\,A BILLEBot prototype swimming excited by two different PWM signals with the frequencies and DC values in parentheses.~\textbf{(e)}\,A FRISSHBot prototype swimming excited by two different PWM signals with the frequencies and DC values in parentheses.~\textbf{(f)}\,\mbox{Electronics-free} \mbox{pneumatic-based} soft robot, shown on the right in \mbox{Fig.\,\ref{FIG01}(l)}, crawling inside a trench. This robot is \mbox{feedback-controlled} using neuromorphic mechanical computation and mechanical tactile and proprioceptive sensors. \label{FIG07}}
\vspace{0.0ex}
\end{figure*}

The \mbox{SMALLBug} design materializes the notion of EI because the mechanical, thermodynamic, and material properties of the driving \mbox{SMA-based} actuator---and its integration with the robot's frame and legs---enable locomotion patterns to emerge without explicit sensing, computation, or complex control logic. For example, when its actuator is excited using a \textit{\mbox{pulse-width} modulation} (PWM) voltage, changing actuation frequency alone switches gaits between crawling (\mbox{$2$\,Hz}), shuffling (\mbox{$10$\,Hz}), and galloping (\mbox{$20$\,Hz}). Notably, different locomotion modes emerge from \mbox{actuator–body–environment} interaction, without the need for active gait control. The \mbox{SMARTI}---the crawler shown in \mbox{Fig.\,\ref{FIG01}(d)}---is composed of two \mbox{SMALLBug} platforms connected in parallel. This configuration is \mbox{$2$D} steerable by simply exciting its two driving actuators with \mbox{phase-shifted} PWM voltages; thereby once more leveraging anisotropic friction for functionality and control. Pushing the \mbox{EI-enabled} approach further, we developed the SPARQ, shown in~\mbox{Fig.\,\ref{FIG01}(e)}, and MiniBug, shown in~\mbox{Fig.\,\ref{FIG01}(f)}. The SPARQ at \mbox{500\,mg} (including its battery) is the lightest autonomous crawler developed to date, and the \mbox{MiniBug} at \mbox{$10$\,mg} is the lightest and smallest crawler with onboard actuation ever reported. A photographic composite of a MiniBug prototype crawling on a flat surface is shown in~\mbox{Fig.\,\ref{FIG07}(b)}. Video footage of six experiments of this type is shown in the accompanying supplementary movie.

\subsection{Anisotropic Drag for Surface Swimming}
\vspace{-0.5ex}
\label{SEC03D}
The \mbox{WaterStrider}---shown in \mbox{Fig.\,\ref{FIG01}(g)}---is a surface swimmer bioinspired by \textit{Gerridae}~\cite{TrygstadCK2023}. Similarly to the crawlers discussed in Section\,\ref{SEC03C}, this robot is driven by two \mbox{SMA-based} bending actuators and leverages \mbox{EI-inspired} methods to function. Specifically, it propels itself forward using anisotropic drag and floats by means of the surface tension of water. Using a \mbox{rowing-resembling} subsemicircular motion pattern, its two swimming appendages rotate rigidly (generating relatively high drag) during the power stroke and compliantly during the reverse stroke (generating relatively low drag). Thus, once more, locomotion emerges from the integration of morphology and materials, and environmental interactions, rather than from complex computation or sensing. In this sense, the \mbox{WaterStrider} \textit{embodies} intelligence, as it achieves functional behaviors---floating, moving, turning---through design choices that leverage the interaction with the environment, thus offloading computational burden to morphology and material properties. Photographic composites of a WaterStrider prototype swimming excited by four different \mbox{open-loop} signals are shown in~\mbox{Fig.\,\ref{FIG07}(c)}. Footage of a test of this type is shown in the supplementary movie.

\subsection{FSI-Based Generation of Traveling Waves for Swimming}
\vspace{-0.5ex}
\label{SEC03E}
The VLEIBot, VLEIBot\textsuperscript{+}, VLEIBot\textsuperscript{++}, and \mbox{BILLEBot} platforms, first presented in~\cite{BlankenshipEK2024,LongwellCR2024,TrygstadCK2025I}, are shown in \mbox{Figs.\,\ref{FIG01}(h)--(j)}. As seen, the first robot is propelled by a single propulsor and the last three robots by two. Each propulsor is composed of two main elements:\,(i)\,an \mbox{SMA-based} bending actuator of the type discussed in Section\,\ref{SEC03C};\,and,\,(ii)\,a passive flexible hydrofoil, which functions as a propulsive fin. As depicted in \mbox{Fig.\,\ref{FIG06}(b)}, when a propulsive fin of the type driving the robots shown in \mbox{Figs.\,\ref{FIG01}(h)--(j)} is flapped by its actuator, a traveling wave is generated through a complex nonlinear mechanism of FSI. Fully understanding how a traveling wave is created through FSI is an extremely difficult task and requires the use of \textit{computational fluid dynamics} (CFD). However, using the methods discussed in~\cite{LighthillMJ1960,LighthillMJ1971,WuTY1971} and video analyses, we can estimate the generated propulsive forces by considering the fin's motion to be fully prescribed and produced by a \textit{virtual~actuator}. This approach is based on the notion of \textit{added mass}, assuming a slender body of the propulsive tail; therefore, it accounts for the reactive (inertial) hydrodynamic forces acting on the swimmer's body, while resistive (viscous) forces are entirely neglected. Despite these limitations, we have consistently used this method for robotic and controller design. The robots discussed in this subsection exemplify EI because their propulsion arises from the physical coupling of SMA-based actuators, flexible fins, and surrounding fluid. Through FSI, traveling waves and, as a result, propulsive forces emerge naturally without requiring complex computation or control. Thus, functionality and efficiency are achieved by exploiting morphology and physical phenomena directly, rather than relying solely on control algorithms. Photographic composites of a BILLEBot prototype swimming excited by two different \mbox{open-loop} signals are shown in~\mbox{Fig.\,\ref{FIG07}(d)}. Video footage of a test of this type is shown in the accompanying supplementary movie.

\subsection{Swimming by \mbox{Tail-Generated} Vortices}
\vspace{-0.5ex}
\label{SEC03F}
\mbox{Fig.\,\ref{FIG01}(k)} shows the old (left) and new (right) \mbox{FRISSHBot} platforms~\cite{TrygstadCK2024,TrygstadCK2025II}. These two swimmers are composed of two plates connected by an \mbox{SMA-based} bending actuator that applies periodic torques---with identical magnitudes and opposite directions---to both plates during operation. The actuation torques induce hydrodynamic reactive torques---generated by aggregated inertial and viscous forces---on the two plates which, as a consequence, produce the thrust required for swimming. Specifically, by design, the front plate functions as an anchor and the rear plate as a caudal fin. A caudal fin generates thrust by flapping laterally in a way that sheds a series of coherent vortices into the surrounding fluid. These resulting vortices, as a whole, can be thought of as a reverse \mbox{von~K\'arm\'an} vortex street, which produces a narrow, \mbox{high-speed} jet directed rearward. By conservation of momentum, the rearward jet induces a forward reactive force on the robot's body, as shown in \mbox{Fig.\,\ref{FIG06}(c)}. The efficiency of this mechanism is closely tied to the Strouhal number, which quantifies the relative importance of vortex spacing, \mbox{tail-beat frequency}, and swimming speed. In~\cite{TriantafyllouMS1995}, it is argued that \mbox{vortex-based} jet formation is the fundamental \mbox{fluid-dynamic} process by which caudal fins propel swimming machines---including fish---efficiently. The \mbox{FRISSHBot} platforms embody intelligence by exploiting hydrodynamic laws through their morphology and interaction with the surrounding fluid. The oscillations of their fins shed vortices and form \mbox{thrust-generating} jets, thereby reducing control complexity. Photographic composites of a FRISSHBot prototype swimming excited by two different \mbox{open-loop} signals are shown in~\mbox{Fig.\,\ref{FIG07}(e)}. Video footage of a test of this type is shown in the accompanying supplementary movie.

\subsection{Earthworm-Inspired Locomotion, Perceptive Skins, and Neuromorphic Mechanical Computation}
\vspace{-0.5ex}
\label{SEC03G}
The pneumatic soft robot on the left in~\mbox{Fig.\,\ref{FIG01}(l)} can locomote inside pipes and trenches by leveraging \mbox{pressure-controlled} anisotropic friction according to the pattern shown in \mbox{Fig.\,\ref{FIG06}(d)}, using as sensors a distributed stretchable artificial skin~\cite{CalderonAA2019}. The pneumatic soft robot on the right in~\mbox{Fig.\,\ref{FIG01}(l)} can locomote inside pipes and trenches driven by an entirely \mbox{electronics-free} feedback controller based on neuromorphic mechanical computation, conceptualized according to the notions in~\cite{McCullochPitts1943}, using \mbox{electronics-free} tactile and proprioceptive sensors~\cite{XuK2020}. These designs exemplify EI by integrating sensing, actuation, and control directly into the robot's body. The use of \mbox{pressure-controlled} anisotropic friction and stretchable skins allows locomotion and perception to emerge from material properties rather than from electronics, control, and computation. Likewise, inspired by EI, the neuromorphic mechanical controller achieves feedback regulation by leveraging the intrinsic mechanical dynamics of the robot instead of using electronics. Together, these features demonstrate how morphology and distributed sensing can enable autonomous functionality through embodiment. This feature is especially important in soft robotics. A photographic sequence of the soft robot on the right in \mbox{Fig.\,\ref{FIG01}(l)} locomoting inside a trench is shown in \mbox{Fig.\,\ref{FIG07}(f)}. Video footage of a test of this type is presented in the accompanying supplementary movie.

\section{Conclusion}
\vspace{-0.5ex}
\label{SEC04}
EI, rooted in physical \mbox{co-design}, challenges the traditional view that intelligence resides solely in centralized---digital or analog---computation. \mbox{Electronics-free} microrobots illustrate how control, sensing, and actuation can emerge from morphology and material response. In contrast to classical robotic systems with isolated modules, EI requires integration. As microrobotics advances toward autonomy under extreme constraints, embodied approaches offer both a theoretical foundation and a practical path forward. Embodied systems are not just artifacts of constrained robotics---they are demonstrations of a deeper principle: that behavior can be designed into matter itself. \mbox{Co-design} is not just a method, but an intellectual approach to research and development.

\bibliographystyle{IEEEtran}
\bibliography{references}
\end{document}